\documentclass[runningheads]{llncs}

\usepackage{eccv}
     
\usepackage{eccvabbrv}
\usepackage{algorithm}
\usepackage{algpseudocode}
\usepackage{enumitem}
\usepackage{graphicx}
\usepackage{booktabs}
\usepackage{wrapfig}

\usepackage{multirow}
\def\domainnetczsl{DomainNet-CZSL}
\def\iwildcamczsl{iWildCam-CZSL}
\usepackage[accsupp]{axessibility}  
\usepackage[dvipsnames]{xcolor}

\usepackage{hyperref}

\usepackage{orcidlink}

\begin{document}

\title{Domain Aware Continual Zero-Shot Learning} 

\titlerunning{Domain Aware Continual Zero-Shot Learning}

\author{Kai Yi\inst{1}\and
Paul Janson\inst{1}\and
Wenxuan Zhang\inst{1}\and
Mohamed Elhoseiny\inst{1}}

\authorrunning{Kai et al.}

\institute{King Abdullah University of Science and Technology (KAUST)\\
\email{\{kai.yi,paul.janson,wenxuan.zhang,mohamed.elhoseiny\}@kaust.edu.sa}}

\maketitle

\begin{abstract}
Modern visual systems have a wide range of potential applications in vision tasks for natural science research, such as aiding in species discovery, monitoring animals in the wild, and so on. However, real-world vision tasks may experience changes in environmental conditions, leading to shifts in how captured images are presented. To address this issue, we introduce Domain-Aware Continual Zero-Shot Learning (DACZSL), a task to recognize images of unseen categories in continuously changing domains. Accordingly, we propose a Domain-Invariant Network (DIN) to learn factorized features for shifting domains and improved textual representation for unseen classes. DIN continually learns a global shared network for domain-invariant and task-invariant features, and per-task private networks for task-specific features. Furthermore, we enhance the dual network with class-wise learnable prompts to improve class-level text representation, thereby improving zero-shot prediction of future unseen classes. To evaluate DACZSL, we introduce two benchmarks, DomainNet-CZSL and iWildCam-CZSL. Our results show that DIN significantly outperforms existing baselines by over 5\% in harmonic accuracy and over 1\% in backward transfer and achieves a new SoTA.
\keywords{Transfer Learning \and Continual Zero-Shot Learning \and Domain Generalization}
\end{abstract}
 
\begin{figure}
    \centering
    \captionsetup{type=figure}
    \includegraphics[width=\textwidth]{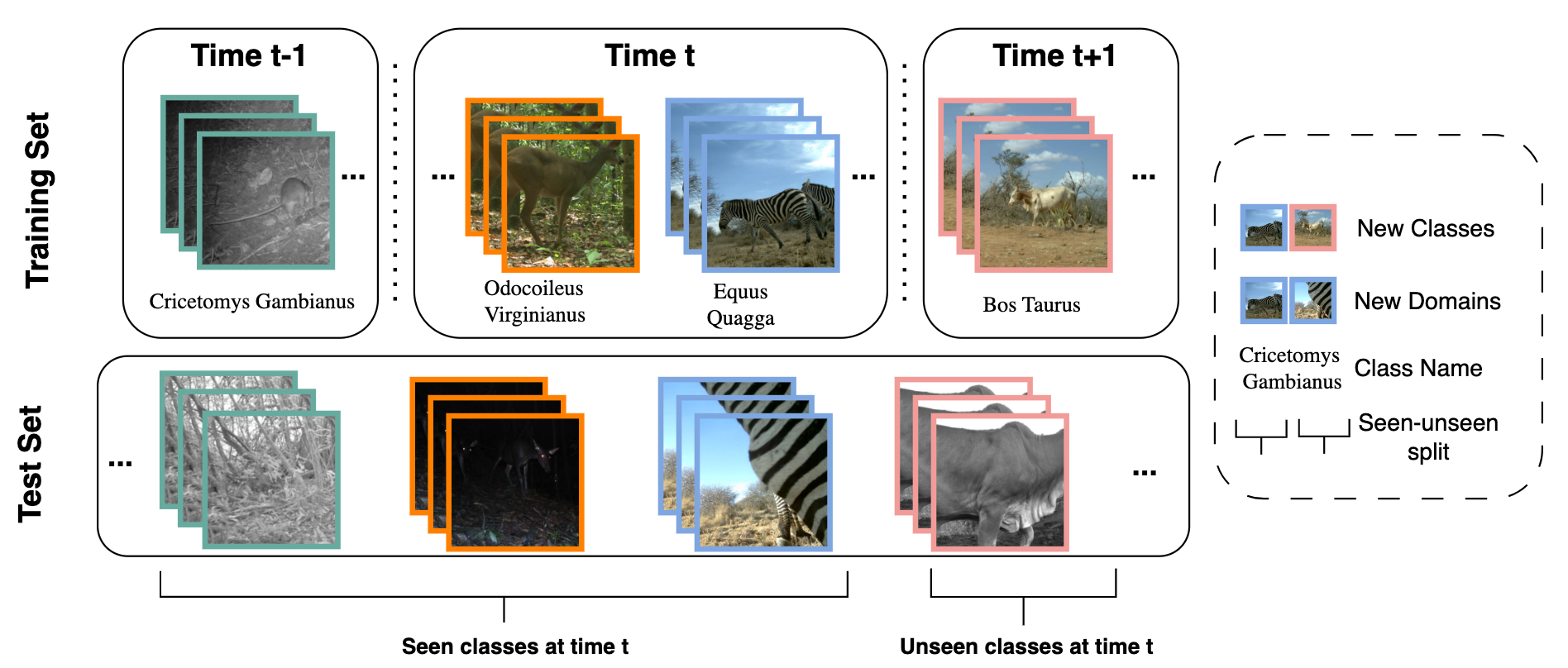}
    \captionof{figure}{\textbf{DACZSL Setting.} Wild images undergo domain shifts due to various factors such as location, time, camera positions, and so on. In DACZSL, at each time $t$, images of new classes are revealed to train the model, and the model is then evaluated on new domains containing both seen classes, \ie, classes revealed up to time $t$, and unseen classes, \ie, classes to be revealed after time $t$.
    }
    \label{fig:teaser-figure}
\end{figure}%

\section{Introduction}\label{section:introduction}
Deep learning-driven computer vision has emerged as a crucial tool for accelerating research in the natural sciences~\cite{beery2018recognition, voina2021biologically}. For example, our planet is estimated to have 8.7 million species, many of which may become extinct even before being documented by humans~\cite{sweetlove2011number}. An efficient classification algorithm can assist in processing the vast amount of images captured in the wild, thereby identifying rare and unknown species.

However, designing visual models for images captured in the wild presents significant challenges~\cite{van2018inaturalist}. For example, wildlife images are often taken by camera traps deployed in various locations. Some animals may frequently visit specific locations, which can potentially cause the model to recognize an animal by the background of the scene rather than the animal itself. Moreover, various factors, such as animal migration, changes in weather and light conditions, and natural disasters, necessitate the system's ability to recognize wildlife against new backgrounds. As shown in~\cref{fig:teaser-figure}, the top row shows an image of an Odocoileus virginianus during the day, while the bottom row illustrates the same species at night. Additionally, models are expected to recognize species that have evolved to thrive in new environments, avoiding the bias towards known species in common environments~\cite{beery2018recognition, beery2021iwildcam}.

To solve such tasks, we seek to develop a deep learning algorithm capable of continually recognizing unseen classes in unseen domains. Prior efforts have partially tackled the challenges. Continual Zero-Shot Learning emphasizes the continuous recognition of unseen categories with a model exclusively trained on seen classes~\cite{farhadi2009describing, elhoseiny2019creativity, han2021contrastive}. However, these models lack a specialized design to encode domain information. Conversely, Domain Generalization aims to apply the knowledge learned from seen domains to unseen target domains~\cite{seo2020learning, gong2019dlow, somavarapu2020frustratingly}. We thus take the domain information into consideration and propose a generalized and unified setting, Domain Aware Continual Zero-Shot Learning (DACZSL).

In this paper, we reexamine the existing works within the DACZSL framework and identify several limitations. First, the text representation techniques employed, such as the Word2Vec representation pre-trained from the Google Corpus~\cite{mikolov2013efficient}, are limited in providing discriminative guidance to recognize very similar and rare classes, such as those related to wild animals as depicted in~\cref{fig:teaser-figure}. Inspired by large-scale vision-language models like CLIP~\cite{radford2021learning} and ALIGN~\cite{Jia2021ScalingUV}, we leverage a Transformer backbone~\cite{vaswani2017attention} to extract text representation of class names. To improve the visual grounding of the class prompts, we further perform visually aligned class-wise prompt learning. Second, training a parameterized classifier over visual features leads to biased recognition towards seen classes when facing unseen classes, particularly across new domains. We address this by implementing an attribute prototypical mapping network and employing contrastive learning on the attribute space. Third, we note that existing methods inadequately consider domain information. To overcome this, we adopt adversarial training to develop domain- and task-invariant features, facilitating knowledge transfer across domains. By integrating these enhancements, we introduce our method, \textit{Domain-Invariant CZSL Network (DIN)}.

To model the setting and evaluate our method, we propose two benchmarks, \domainnetczsl and \iwildcamczsl, by splitting the DomainNet and iWildCam datasets, respectively. We demonstrate superior experimental results of our method on these two benchmarks compared to all baseline models. Furthermore, we conduct ablation studies to assess the impact of each methodological choice. Our \textbf{contributions} can be summarized as follows:
\begin{itemize}
\item We introduce Domain Aware Continual Zero-Shot Learning (DACZSL), which is a novel problem and a step towards the real-world challenge of continuously understanding unseen classes within unseen domains. Our work is the first to investigate the transferability of knowledge in scenarios characterized by both domain and class.
\item We develop a novel DACZSL method, denoted as DIN. DIN introduces informative and visually grounded text representations through an improved text extractor and prompt learning and learns domain- and task-invariant visual features by adversarial training.
\item We propose two benchmarks, \domainnetczsl and \iwildcamczsl, that continually reveal data from new classes and new domains. Through extensive experimentation, we demonstrate that DIN significantly improves performance compared with state-of-the-art methods.
\end{itemize}

\section{Related Works}\label{relworks}
\subsection{Continual Zero-Shot Learning}
Early works in Continual Zero-Shot Learning (CZSL) focus on defining the problem and formulating the setting~\cite{chaudhry2018efficient,wei2020lifelong,skorokhodov2021class}. A-GEM~\cite{chaudhry2018efficient} is the pioneering study to consider continual zero-shot performance, reporting zero-shot accuracy for future tasks in a continual learning stream. Wei et al.\cite{wei2020lifelong} proposed a variant of the CZSL setting termed lifelong ZSL, where the model learns from all the seen classes from various datasets sequentially in a continual learning setting before being evaluated on all the unseen classes. Skorokhodov et al.\cite{skorokhodov2021class} proposed a generalized CZSL setting that expanded the classification space to include all classes from all tasks, with the model expected to continuously recognize more classes in this space. Following Skorokhodov et al.\cite{skorokhodov2021class}, most CZSL methods strive for enhanced performance on extensive feature-level attribute-based datasets. Gautam et al.\cite{gautam2020generalized} proposed a replay-based Variational Autoencoder (VAE) approach to improve the recognition of unseen classes. Ghosh et al.\cite{ghosh2021dynamic} used stacked VAEs with the generative replay of seen samples to mitigate forgetting. Zhang et al.\cite{cgzsl} advanced the field by employing a generative random walk to improve discrimination between unseen classes.

\subsection{Zero-Shot Learning with Large-Scale Pretraining}
In early zero-shot learning, image features are often extracted by ResNet101~\cite{he2016deep} pre-trained on the ImageNet-1K dataset~\cite{russakovsky2015imagenet} for attribute-based datasets, and VGG~\cite{simonyan2014very} for text-based datasets. The semantic class features are processed with GloVe~\cite{pennington2014glove} or Skip-Gram~\cite{mikolov2013distributed}. However, the low-dimensional text features from naive GloVe or Skip-Gram often struggled to distinguish classes well in ZSL. Additionally, training visual and semantic encoders in isolation may not capture the complex vision-language relationships accurately. Recently, pre-trained models from large vision and language pairs such as CLIP~\cite{radford2021learning} and ALIGN~\cite{Jia2021ScalingUV} have shown promising zero-shot learning ability. Specifically, CLIP text and image encoders were collaboratively pre-trained on over 400 million vision-language pairs. Building upon CLIP, recent studies such as CoOp~\cite{zhou2021learning}, CoCoOp~\cite{zhou2022conditional}, and CPL~\cite{he2022cpl} have further developed prompt learning strategies to refine semantic representations. Our method uses pre-trained CLIP and adopts the contrastive prototypical classification design to optimize the attribute prototypical network~\cite{xu2020attribute}.
  
\section{Domain-Aware Continual Zero-Shot Learning}
\subsection{Problem formulation}
In this section, we formally introduce Domain-Aware Continual Zero-Shot Learning. 
Here we start by introducing notations of domain and class. 

Let $\mathcal{H} = \{(X, Y, A)\}$ be the set of data, where $X$, $Y$, and $A$ represent the set of images, their corresponding label set, and the semantic description set, respectively. We construct a squence of $T$ tasks by split $\mathcal{H}$ into $T$ subsets $\{\mathcal{H}_1, \dots, \mathcal{H}_T\}$ with each $\mathcal{H}_i$ has disjoint label space. Following~\cite{skorokhodov2021class}, at task $t$, the model is trained on classes from $ \mathcal{H}_t$, and evaluated on both seen classes from $\{\mathcal{H}_1, \dots, \mathcal{H}_t\}$ and unseen classes from $\{\mathcal{H}_{t+1}, \dots, \mathcal{H}_T\}$.

Let $\Delta_\mathcal{D} = \{\mathcal{D}_i\}_{i=1}^N$ be the set containing  $N$ different domains. We follow \cite{mancini2020towards} to define $\Delta_\mathcal{D}^\text{source} =\{\mathcal{D}_{s_i}\}_{i=1}^{N^s}$ as the set containing $N^s$ source domains  and  $\Delta_\mathcal{D}^\text{target} =\{\mathcal{D}_{t_i}\}_{i=1}^{N^t}$ as the set containing $N^t$ target domains. Assume $N = N^s+N^t$ and thus $\Delta_\mathcal{D} = \Delta_\mathcal{D}^\text{source} \cup \Delta_\mathcal{D}^\text{target}$. The training is always performed on the source domains and the evaluation is on the target domains. 

Combining class and domain notation, at each task $t$, the model is trained on $\mathcal{H}^\text{source}_t = \{X_t, Y_t, A_t | (X_t, Y_t) \in \mathcal{D}^{\text{source}}_t\cap \mathcal{H}_t \}$, and then evaluated with all the classes from target domain $\mathcal{H}^\text{target}_t = \{X_i, Y_i, A_i | (X_i, Y_i) \in \mathcal{D}^{\text{target}}_t \cap \mathcal{H}_t \}_{i=1}^T$. We further assume that the task identifier is known to the model.

\subsection{Source Domain Assumption}
At each task $t$, when we have uniform access to training data from all source domains, \ie, $\mathcal{D}^{\text{source}}_t = \cup \{\mathcal{D}_{s_i}\}_{i=1}^{N^s}$, we denote this setting as the \textit{Uniform DACZSL}. However, this is not always the case in the real world. For example, in Fig.~\ref{fig:teaser-figure}, we cannot see all the animals from a single camera trap, and some animals, such as Cricetomys gambianus, are usually active under specific light conditions. Therefore, we further introduce the \textit{Non-uniform DACZSL}. In this setting, for each task, we randomly remove one domain, $\mathcal{D}_{s_r}$, from the source domains, where $s_r$ is an index selected at random. Thus, in this setting, $\mathcal{D}^{\text{source}}_t = \cup \{\mathcal{D}_{s_i}\}_{i=1, s_i\ne s_r}^{N^s}$.
Non-uniform DACZSL presents a greater challenge, as it requires our model to alleviate the bias towards invisible domains.  The invisible domains include both the unseen target domains and randomly removed source domains. 

\begin{figure}[t]         
    \centering
    \includegraphics[width=0.95\textwidth]{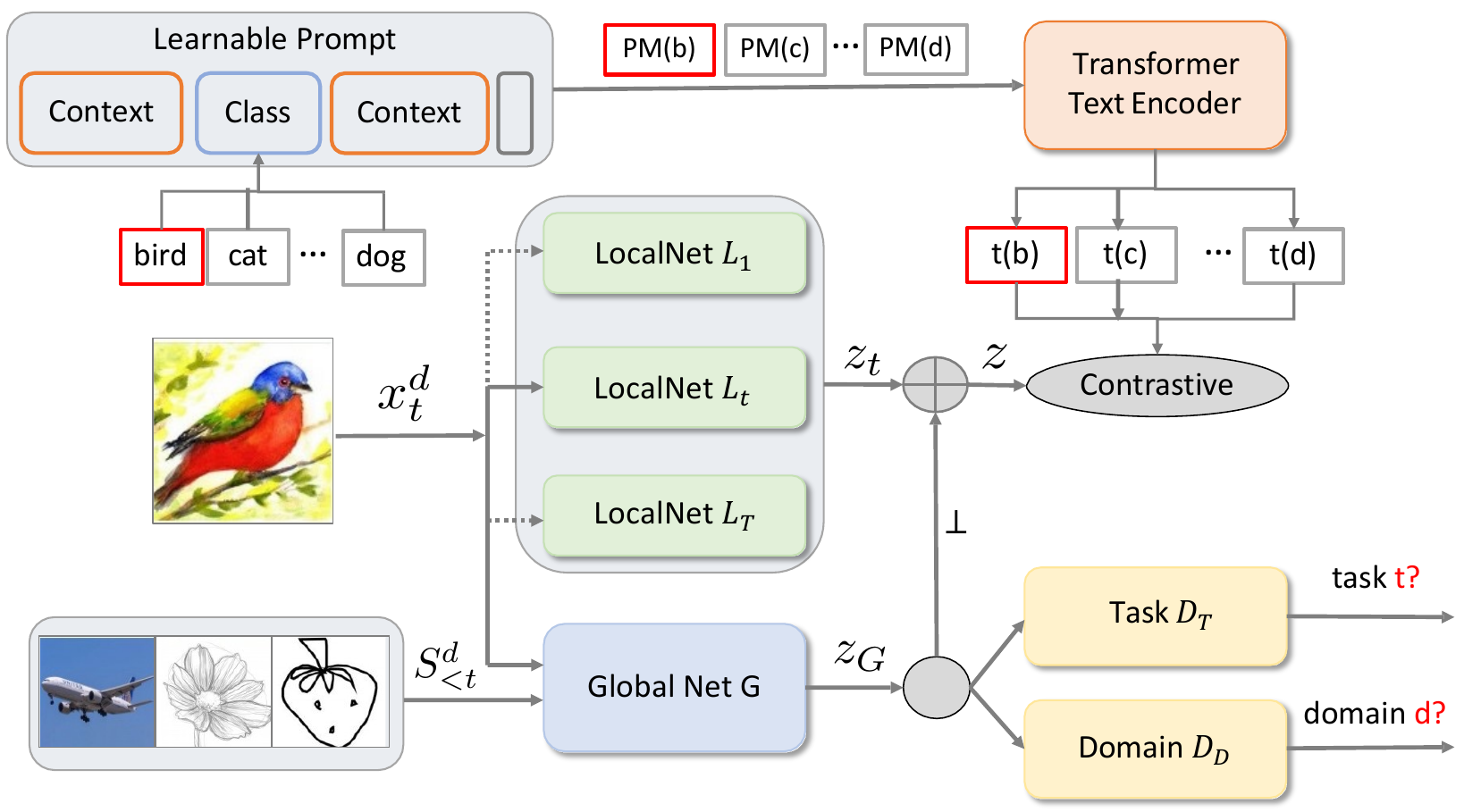}
    \caption{DIN involves two key steps. First, we concatenate class-wise learnable prompts to each class token embedding. These prompts are learned by minimizing the contrastive loss between the text feature $t(\texttt{CLASS})$ and the global visual feature $z_G$ from the global network $G$. Next, we employ adversarial training to refine the global feature $z_G$, and implement disentanglement techniques for the local feature $z_t$. This approach ensures that the global network encodes domain- and task-invariant information, while the local network processes task-specific information.}    
    \label{fig:main_structure}
\end{figure}

\section{Method}
\label{sec:method}
In this section, we propose our Domain Invariant Network to address the domain-aware continual zero-shot learning problem. To enhance text representation, we introduce visually guided, domain-invariant prompt learning as detailed in~\cref{method:class-wise-learnable-prompt}. To improve domain discrimination capability, we propose learning factorized visual features through adversarial training, as detailed in~\cref{method:Adversarial-knowledge-disentanglement}. Finally, we align visual and text features using contrastive learning, as described in~\cref{method:contrastive-fine-tuning}.

\subsection{Class-wise Learnable Prompts}\label{method:class-wise-learnable-prompt}
As mentioned in~\cref{section:introduction} and illustrated in~\cref{fig:teaser-figure}, traditional texture attributes or word embeddings may struggle in scenarios like distinguishing between rare class names. To address such issues, we enhance text embeddings with additional parameters through prompt learning, guiding the learning by visual content from multiple domains.

Prompt learning has been extensively studied in natural language processing~\cite{autopromptemnlp20,jiang2020can,zhong2021factual,le2021many}, and zero-shot learning~\cite{zhou2021learning}. Following the standard process, we propose encoding both class and domain information into the learnable prompts, as illustrated in the top-left part of~\cref{fig:main_structure}. Specifically, we concatenate learnable prompts $[W_i]_{i=1}^k$ to the semantic embedding of each class name $c$, and obtain the learnable textual embedding PM($c$) for class $c$ as 
\begin{equation}
    \text{PM}(c) = [W_1,\dots,W_{k'}, \text{EM}(c),  \dots, W_k, \texttt{EOS}],
\end{equation}
where $ \text{EM}(c)$ is the embedding of the class name, \texttt{EOS} is the end-of-sentence token, and $W_i$, $1\leq i\leq k$, are the learnable embeddings. Here, $k'$ indicates the insertion position of the class name embedding. We consider two variants of where to add the class token, either in the middle, \textit{i.e.,} $k' < k$ or at the end of the prompt, \textit{i.e.,} $k'= k$. The difference between these variants is minor, as shown in our experimental results in~\cref{fig:prompts}. We report the final results with the latter variant, \textit{i.e.,} placing the class prompt embedding at the end.

To train the class-wise learnable prompt with multi-domain and multi-class information, we construct a balanced prompt learning set $P_t$ by randomly sampling $K$ samples per domain and per class from the training set of the current task $H^\text{source}_t$. We perform prompt learning using the contrastive loss between the text encoder and the global net. The global net here produces domain-invariant features, which we will introduce later. Specifically, the loss can be written as 
\begin{equation}\label{eq:promptlearning}
     \mathcal{L}_{\text{prompt}}=- \sum_{x_i \in P_t}\log \frac{\exp \left(\operatorname{r}\left({t}_{i}, {z_G}\right) / \tau\right)}{\sum_{k=1}^{N^c_t} \exp \left(\operatorname{r}\left({t}_{k}, {z}\right) / \tau\right)},
\end{equation}
where $x_i$ is sampled from the balanced set $P_t$ of task $t$ and belongs to class $i$, $t_i = T( \text{PM}(c_i))$ is the text feature of class $i$ with learnable prompt by text encoder $T$, $z_G = G(x_i)$ is the global visual feature by global net $G$,  $N^c_t$ is the number of classes from the current balanced set $P_t$, $r$ is the similarity measurement, and $\tau$ is the temperature hyper-parameter.

Intuitively, the domain-invariant visual features help the learnable prompt to capture the concept represented by multiple domains. This improves discrimination within rare concepts and could mitigate representative bias towards seen domains during evaluation.

\subsection{Adversarial Knowledge Disentanglement}
\label{method:Adversarial-knowledge-disentanglement}
Besides improving discrimination between text features, we further factorize visual content into domain-invariant knowledge from domain- and task-specific information. We illustrate this process at the left bottom in~\cref{fig:main_structure}. Let $x_t^d$ represent an image from domain $d$ with label $c$ in task $t$. We process this image through Local Nets $L_t(\cdot)$ and Global Net $G(\cdot)$ to obtain local feature $z_t$ and global feature $z_G$. Here, $L_t(\cdot)$ is instantiated for each task $t \in \{1,\cdots,T\}$.

To make the global feature $z_G$ domain and task invariant, we propose an adversarial training scheme, akin to the intuition behind GAN~\cite{goodfellow2014generative} and ACL~\cite{ebrahimi2020adversarial}, but without using fake data. Similar to the GAN structure, we have a shared network to extract features and discriminators to predict the task and domain of the features. We iteratively train the shared network and discriminator. We design two discriminators, \textit{i.e.,} the task discriminator $D_T$, evaluating which task the generated sample is from, and the domain discriminator $D_D$, classifying the domain of the current sample. During the training of the shared net, we expect the shared features from different domains and tasks cannot be differentiated by the discriminator. The loss of the shared net is 
\begin{equation}\label{eq_adv}
    \mathcal{L}_{\text{adv}}= \sum_{x_t^d\in \mathcal{H}_t^{\text{source}}} \left[ \sum_{i_t=1}^T i_t \log (D_T(G(x_t^d)))+\sum_{i_d=1}^{D}  i_d\log (D_D(G(x_t^d))) \right].
\end{equation}
Here we assign fake task labels $i_t$ for $i_t =1, \dots,T$ and fake domain labels $i_d$ for $i_d=1,\dots,N$ to every input image $x_t^d$, where $T$ is the total number of tasks and $D$ is the total number of domains. $G(x_t^d)$ produces the shared features, and $D_T(G(x_t^d))$ and $D_D(G(x_t^d))$ make the task and domain predictions. We minimize the entropy between task and domain predictions for each of its fake labels.

To train the discriminator, we use the correct label for each data point. For a data point $x_t^d$ from task $t$ and domain $d$, the loss is 
\begin{equation}\label{eq:discriminator}
    \mathcal{L}_{\text{disc}}=  \sum_{x_t^d\in \mathcal{H}_t^{\text{source}}} \left[ t \log (D_T(G(x_t^d)))+ d\log (D_D(G(x_t^d))) \right].
\end{equation}
Here we minimize the entropy between the domain logits and task logits of shared features $G(x_t^d)$ to train the discriminators $D_T$ and $D_D$. Unlike standard adversarial training in ACL and GAN, we don't train discriminators to identify real or fake samples. Instead, for every task $t$ and domain $d$, we view the real samples from other domains or tasks as ``fake'' samples of the current domain and task, and optimize the shared net to fool the discriminators by them. If the discriminators fail to predict the domain and tasks, this means the generator, \textit{i.e.,} shared net, is able to produce features of domain $d$ and task $t$ that are as similar as features from other tasks and domains. As a result, these features are domain and task invariant.

For every task, we also design a local net to capture task-specific information. To encourage feature factorization and improve orthogonality between global feature $z_G$ and local feature $z_t$, we apply the disentangled loss between them as
\begin{equation}
    \mathcal{L}_{\text{disen}} = \sum_{x_t^d \in \mathcal{H}_t^{\text{source}}} \|\langle z_t, z_G\rangle\|^2 = \sum_{x_t^d \in \mathcal{H}_t^{\text{source}}} \|\langle L_t(x_t^d), G(x_t^d)\rangle\|^2 .
    \label{eq_disent}
\end{equation} 
Finally, we may optionally store a small set of samples from previously seen tasks, denoted as $S^d_{<t}$, where $ (x^d,y^d)\in \mathcal{D}^\text{source}_{<t}$.

\subsection{Visual-Text Alignment}
\label{method:contrastive-fine-tuning} 
We use contrastive loss, as shown in Eqn.~\ref{eqn:contrast}, to encourage the features from the local nets and the global net, denoted by ${z}=[z_t, z_G]$, to be closer to the label text feature representation ${t}_i$ of the same class, and farther from those of different classes. $t_i = T( \text{PM}(c_i))$ is the text representation of class $i$ by the text encoder $T$.
\begin{equation}\label{eqn:contrast}
    \mathcal{L}_{\text{cont}}=-\log \frac{\exp \left(\operatorname{r}\left({t}_{i}, {z}\right) / \tau\right)}{\sum_{k=1}^{N_{BS}} \exp \left(\operatorname{r}\left({t}_{k}, {z}\right) / \tau\right)},
\end{equation}
Here, $N_{BS}$ represents the number of samples in the current batch, and $\tau$ is the temperature parameter. The function $\operatorname{r}\left(\cdot,\cdot\right)$ computes the cosine similarity between the two input feature representations.

\subsection{Training Pipline}
\begin{algorithm}[t]
\begin{algorithmic}[1]
\For{ task $t$ in $1,2,\dots,T$ }
    \State Receive training set $H_t^{\text{source}}$
    \State Construct balanced prompt learning set $P_t$
    \State Freeze $G, L_{1:T}, D_D, D_T$,  and learn prompts $[W_i]_{i=1}^k$ by in $\mathcal{L}_{\text{prompt}}$~\cref{eq:promptlearning}
    \State Freeze the $[W_i]_{i=1}^k, D_D, D_T$ , and train Local Net $L_t$ and Global Net $G$ by  $ \mathcal{L}_{\text{DIN}} $ in~\cref{eq:din}
    \While{Discriminator Training}
    \State Freeze $[W_i]_{i=1}^k,G, L_{1:T}$,  and train discriminators $D_D, D_T$ by $\mathcal{L}_{\text{disc}}$ in~\cref{eq:discriminator}
    \EndWhile
    \State (Optional) Update replay buffer $S^d_{<t}$
    \EndFor
    \end{algorithmic}
\caption{ Domain-Invariant Network for DACZSL}%
\label{alg:din}%
\end{algorithm}

During the training, we update the prompt, the main networks (including the global and local networks and the text encoder), and the discriminators separately, as shown in~\cref{alg:din}. Upon receiving data, we first perform prompt learning (lines 3-4), which provides a quality text prompt for later visual-text alignment. Then, we update the main network (line 5). The loss function for updating the main network is as follows,
\begin{equation}  \label{eq:din}
    \mathcal{L}_{\text{DIN}} = \lambda_1 \mathcal{L}_{\text{disen}} + \lambda_2 \mathcal{L}_{\text{adv}} + \lambda_3 \mathcal{L}_{\text{cont}},
\end{equation}
where $\lambda_1$, $\lambda_2$, and $\lambda_3$ are hyperparameters that control the influence of different loss components. The disentangled loss $\mathcal{L}{\text{disen}}$ encourages features in the global network to be independent of any task-variant local net, while the adversarial loss $\mathcal{L}{\text{adv}}$ makes the knowledge in the global net task and domain invariant. The contrastive loss $\mathcal{L}_{\text{cont}}$ is used for class-level prompt learning. Finally, we train the discriminator for a few steps (lines 6-8) to make it distinguishable for the current task and domain. For more details and the algorithm, please refer to the supplementary.

\section{Experiments}\label{experiments}
\subsection{Benchmarks}\label{benchmarks}
To formulate the DACZSL setting, we introduce two benchmarks: DomainNet-CZSL and iWildCam-CZSL.

\noindent\textbf{DomainNet-CZSL.} We build DomainNet-CZSL based on DomainNet~\cite{peng2019moment}, originally designed for multi-source domain adaptation. It contains approximately 0.6 million images from 345 classes and 6 domains. We use the noise-reduced version of the dataset with 4 domains \cite{saito2019semi}.
For DACZSL, we randomly split DomainNet into 8 tasks, each containing 43 classes. We removed the last class to ensure each task contains an equal number of classes for stable training.
The original dataset does not provide attribute information for each class. We use word2vec embeddings~\cite{mikolov2013efficient} extracted from the Google News corpus to obtain a 300-dimensional representation of each class and perform $L_2$ normalization afterward.

\noindent\textbf{iWildCam-CZSL.} To approach real-world applications of DACZSL more closely, we utilize the iWildCam dataset~\cite{beery2021iwildcam} from the Wilds benchmark~\cite{pmlr-v139-koh21a} to construct iWildCam-CZSL. The camera trap locations are treated as domains. We use the in-domain split as the source domain and the out-of-domain split as the target domain.
Source domains consist of 243 camera trap locations, and target domains consist of 48 camera trap locations. We split the 312 classes into 7 tasks, each comprising 26 classes, for DACZSL.

\subsection{Implementation Details}\label{details}
\textbf{Evaluation Metrics.}
We adopt evaluation metrics from prior studies \cite{chaudhry2018efficient,skorokhodov2021class,cgzsl} for DACZSL, including last seen accuracy (LS), mean seen accuracy (mS), mean unseen accuracy (mU), the harmonic mean of seen and unseen accuracy (mH), and the forgetting rate (BWT) across task time steps. Detailed definitions are available in the supplementary material.

\noindent\textbf{Baselines.}\label{baselines}
To evaluate our method's performance, we establish strong baselines by adapting SoTA methods from related fields. We include CuMix~\cite{mancini2020towards} and its variants from domain generalization tasks, continual zero-shot learning methods like CNZSL~\cite{skorokhodov2021class}, BDCZSL~\cite{9879919}, and IGCZSL~\cite{cgzsl}, as well as results from frozen and finetuned CLIP~\cite{radford2021learning}. Moreover, we apply the text encoder from frozen CLIP across all baselines.

\noindent\textbf{Backbone, Training, and Hyperparameters.}\label{details}
For all experiments, we employ a ResNet-50~\cite{he2016deep} backbone for visual input. Both the domain discriminator and the task discriminator consist of 3-layer MLPs. Optimization is performed using AdamW~\cite{loshchilov2017decoupled}. While we adhere to default hyperparameters for baselines, we adjust key parameters based on validation set performance. Comprehensive implementation details are provided in the supplementary material.

\begin{table}[t]\centering
\resizebox{\textwidth}{!}{
\begin{tabular}{l|ccc|ccc|ccc|ccc}
\toprule 
Target Domain& \multicolumn{3}{|c|}{Sketch} & \multicolumn{3}{|c|}{Clipart} & \multicolumn{3}{|c|}{Painting} & \multicolumn{3}{c}{Avg.}\\ \cmidrule(l){1-4} \cmidrule(l){5-7} \cmidrule(l){8-10} \cmidrule(l){11-13}
 {Method} & LS$\uparrow$ & mH$\uparrow$ & BWT$\uparrow$ & LS$\uparrow$ & mH$\uparrow$ & BWT$\uparrow$ & LS$\uparrow$ & mH$\uparrow$ & BWT$\uparrow$ & LS$\uparrow$ & mH$\uparrow$ & BWT$\uparrow$\\  
\midrule
CuMix~\cite{mancini2020towards} & 6.97& 3.58  & 0.15 & 6.32  &  3.11  & 0.24  & 6.80  &  3.47  & 0.09  & 6.70  & 3.39  & 0.16 \\   
CuMix w/o Mixup & 2.67  & 1.92  & -0.03  & 2.36  & 1.63 & 0.02  & 2.80  & 2.01  & -0.05  & 2.61  & 1.85 & -0.02 \\
CuMix + tf & 9.52 & 4.06  & 0.32  & 10.25  & 4.09  & 0.33  & 9.57  & 4.69  & 0.06  & 9.78   & 4.28  & 0.24  \\ \midrule
CNZSL~\cite{skorokhodov2021class} & 6.68  & 2.52  & -0.66  & 6.40  & 2.79  & -0.71  & 7.82  & 2.09  & -0.27  & 6.97 & 2.47  & -0.55 \\
CNZSL w/o CN & 7.16  & 2.67 & -0.75  & 7.07  & 2.92  & -0.72  & 8.36  & 2.15 & -0.35  & 7.53  & 2.58  & -0.61  \\
CNZSL + tf & 45.74 & 31.75  & 1.00  & 54.19  & 37.33  & \textbf{1.17 } & 59.15  & 40.37  & 0.96 & 53.03 & 36.48  & 1.04  \\ \midrule
BDCZSL~\cite{9879919}& 15.62 & 6.98 & -15.54 & 16.61 & 7.60 & -16.45 & 15.96 & 7.69 & -17.54 & 16.06 & 7.42 & -16.51\\
BDCZSL + tf& 33.20 & 28.51 & -0.46 & 39.69 & 36.39 & -4.13 & 54.95 & 51.82 & -2.89 & 42.61 & 38.91 & -2.49\\ \midrule
IGCZSL~\cite{cgzsl}& 33.42 & 12.81 & -28.01 & 36.93 & 14.46 & -32.83 & 39.16 & 15.66 & -30.85 & 36.51 & 14.31 & -30.56\\
IGCZSL + tf & 26.36 & 23.14 & \textbf{1.49} & 29.21 & 27.27 & 0.73 & 54.15 & 51.51 & 1.03 & 36.57 & 33.97 & 1.08\\ \midrule
CLIP-Frozen~\cite{radford2021learning} & 21.39 & 20.83 & 0.00 & 27.73 & 27.46 & 0.00 & 63.06 & 62.35 & 0.00 & 37.39 & 36.88 & 0.00 \\
CLIP-Finetune & 60.17 & 60.23 & 0.60 & 67.04 & 67.18 & 0.41 & 61.51 & 62.00 & 0.29 & 62.91 & 63.14 & 0.43  \\ \midrule
\textbf{DIN} & \textbf{74.53}  & \textbf{63.78} & 0.92  & \textbf{81.31 } & \textbf{71.24 } & 1.12 & \textbf{76.00} & {{\textbf{76.10}}}  & \textbf{1.82}  & \textbf{77.28}  &   \textbf{70.37 } & \textbf{1.29}  \\  
\bottomrule\end{tabular}}
\caption{Three experimental sets on \domainnetczsl{} under \textit{uniform} settings, designating \texttt{Sketch}, \texttt{Clipart}, and \texttt{Painting} as the target domains, respectively. The suffix "+ tf" indicates the utilization of the {CLIP}~\cite{radford2021learning} pre-trained text Transformer for feature extraction.}\label{uniform-daczsl-results}
\vspace{-6.0mm}
\end{table}

\begin{table}[t]\centering
\resizebox{1\textwidth}{!}{%
\begin{tabular}{l|ccc|ccc|ccc|ccc}
\toprule 
\multirow{2}{*}{Method} & \multicolumn{3}{|c|}{Sketch} & \multicolumn{3}{|c|}{Clipart} & \multicolumn{3}{|c|}{Painting} & \multicolumn{3}{c}{Avg.}\\ \cmidrule(l){2-4} \cmidrule(l){5-7} \cmidrule(l){8-10} \cmidrule(l){11-13}
 & LS$\uparrow$ & mH$\uparrow$ & BWT$\uparrow$ & LS$\uparrow$ & mH$\uparrow$ & BWT$\uparrow$ & LS$\uparrow$ & mH$\uparrow$ & BWT$\uparrow$ & LS$\uparrow$ & mH$\uparrow$ & BWT$\uparrow$\\  
\midrule
CuMix~\cite{mancini2020towards} & 5.39 & 3.08 & 0.20 & 5.81 & 2.95 & -0.15 & 5.35 & 2.98 & 0.22 & 5.52 & 3.00 & 0.09\\   
CuMix w/o Mixup & 2.61 & 1.88 & 0.00 & 2.34 & 1.61 & -0.03 & 2.72 & 2.72 & -0.08 & 2.58 & 1.80 & -0.04\\
CuMix + tf & 6.94 & 3.91 & 0.13 & 9.30 & 4.06 & 0.28 & 8.60 & 4.51 & -0.41 & 8.28 & 4.16 & 0.00\\ \midrule
CNZSL~\cite{skorokhodov2021class} & 6.37 & 2.36 & -0.45 & 5.94 & 2.75 & -0.46 & 6.89 & 1.98 & -0.48 & 6.40 & 2.36 & -0.46\\
CNZSL w/o CN & 3.22 & 1.42 & -0.65 & 3.92 & 1.63 & -0.10 & 3.87 & 1.61 & -0.53 & 3.67 & 1.55 & -0.43 \\
CNZSL + tf & 41.20 &  28.59 &  0.89 &  48.43 &  28.73 & {2.93} &  52.39 &  34.5 &1.26 &  47.34 & 30.61 &{1.69} \\ \midrule
BDCZSL~\cite{9879919}& 10.34 & 4.98 & -7.50 & 8.62 & 4.96 & -8.80 & 7.88 & 4.86 & -7.56 & 8.95 & 4.93 & -7.95\\
BDCZSL + tf& 19.15 & 16.93 & 4.68 & 25.59 & 24.14 & 2.15 & 42.09 & 40.39 & 1.28 & 28.94 & 27.15 & 2.70\\ \midrule
IGCZSL~\cite{cgzsl}& 28.01 & 9.71 & -11.80 & 28.86 & 10.78 & -12.49 & 26.13 & 10.33 & -12.12 & 27.66 & 10.27 & -12.14\\
IGCZSL + tf & 26.63 & 23.42 & \textbf{6.00} & 30.68 & 28.56 & \textbf{5.65} & 45.73 & 44.07 & 0.52 & 34.35 & 32.02 & \textbf{4.06}\\ \midrule
CLIP-Frozen~\cite{radford2021learning} & 21.39 & 20.83 & 0.00 & 27.73 & 27.46 & 0.00 & 63.06 & 62.35 & 0.00 & 37.39 & 36.88 & 0.00 \\
CLIP-Finetune & 58.98 & 60.01 & 0.25 & 65.77 & 66.49 & 0.00 & 61.56 & 61.33 & 1.02 & 62.10 & 62.61 & 0.42 \\ \midrule
\textbf{DIN} & \textbf{ 70.99} & \textbf{ 60.14} & 0.81 &  \textbf{76.01} & \textbf{ 69.18} &  0.65 &  \textbf{71.19} &  \textbf{71.96} & \textbf{1.55} & \textbf{ 72.73} &   \textbf{ 67.09} &  1.00 \\  
\bottomrule
\end{tabular}
}
\caption{Three experimental groups conducted on \domainnetczsl{} under \textit{non-uniform} settings, targeting \texttt{Sketch}, \texttt{Clipart}, and \texttt{Painting} as respective domains. The notation "+ tf" signifies the employment of the {CLIP}~\cite{radford2021learning} pre-trained text Transformer for feature extraction.}\label{non-uniform-daczsl-results}
\vspace{-6.0mm}
\end{table}

\begin{figure}[t]
\centering
\begin{minipage}[b]{0.49\textwidth}
\centering\small
\resizebox{\columnwidth}{!}{%
\begin{tabular}{@{}lccc@{}}
\toprule
Method     & \multicolumn{1}{c}{LS$\uparrow$}      & \multicolumn{1}{c}{mH$\uparrow$}      & \multicolumn{1}{c}{BWT$\uparrow$} \\ \midrule
CuMix + tf~\cite{mancini2020towards} & 11.56                       & {\color[HTML]{363A3D} 3.24} & -0.18                   \\
CNZSL + tf~\cite{skorokhodov2021class} & 34.34                       & 19.98                       & -24.35                  \\
BDCZSL + tf~\cite{9879919} &25.81&7.56&-6.51\\ 
IGCZSL + tf~\cite{cgzsl}&32.30& 3.29&-21.35\\ \midrule
CLIP-Frozen~\cite{radford2021learning} & {\color[HTML]{363A3D} 9.21} & {\color[HTML]{363A3D} 8.27} & \textbf{0.00}                    \\
CLIP-Finetune~\cite{radford2021learning}  & 36.01                       & 14.22                       & -42.38                  \\ \midrule
\textbf{DIN} & \textbf{42.62} & \textbf{36.47} & -13.48 \\
\bottomrule
\end{tabular}%
}
\captionof{table}{Results on the iWildCam-CZSL dataset demonstrate that our method, DIN, achieves significantly higher harmonic mean accuracy compared to other methods.}
\label{tab:iwild-cam-results}
\end{minipage}
\hfill
\begin{minipage}[b]{0.47\textwidth}
\centering
\includegraphics[trim = 0 0 0 0, clip, width=0.95\textwidth]{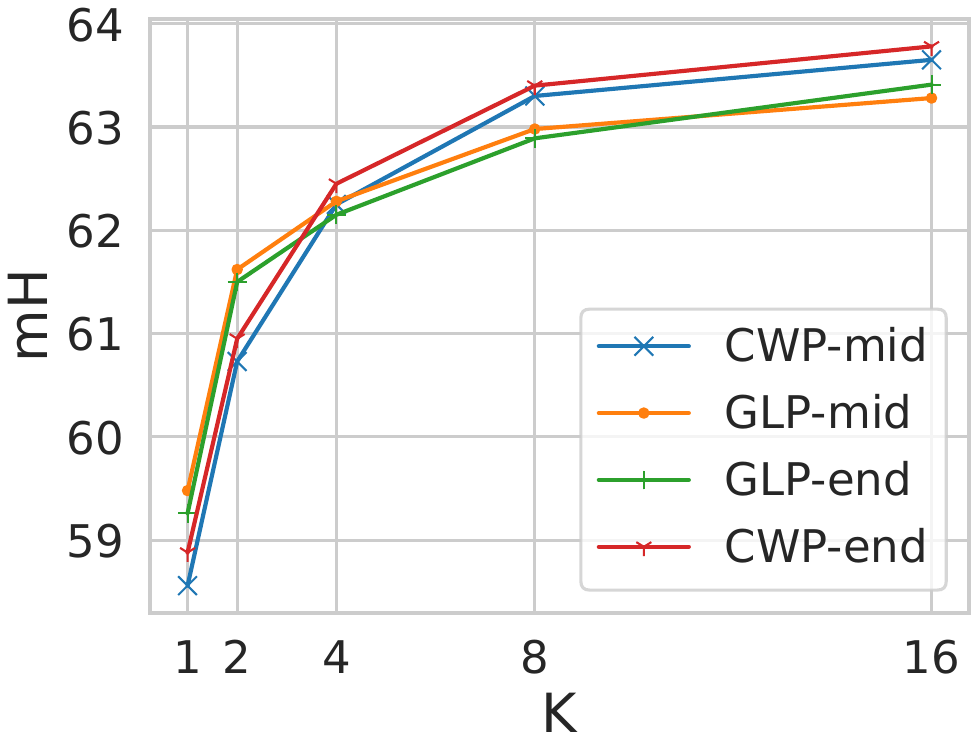}
\vspace{-2mm}
\caption{Harmonic mean of seen and unseen accuracy in experiments on target domain \texttt{sketch}. Curves represent DIN with varying $K$ numbers of per-class per-domain data in prompt learning.}
\label{fig:prompts}
\end{minipage}
\vspace{-2mm}
\end{figure}

\subsection{Experimental Results}
We present results for uniform DACZSL in Table~\ref{uniform-daczsl-results} and non-uniform DACZSL in Table~\ref{non-uniform-daczsl-results}. In the uniform setting, we conduct three groups of experiments that use \texttt{Sketch}, \texttt{Clipart}, and \texttt{Painting} as the target domain, with the remaining three domains serving as source domains, respectively. In the non-uniform setting, we randomly remove one of the three source domains from its counterpart uniform experiment, resulting in two source domains and one target domain in each group of non-uniform experiments.

In the uniform setting, our approach significantly outperforms SoTA methods in related fields, such as CNZSL~\cite{skorokhodov2021class}, BDZSL~\cite{9879919}, IGCZSL~\cite{cgzsl}, and CuMix~\cite{mancini2020towards}, along with their variants. Our contrastive prototypical classifier in DACZSL can be dynamically modified with new class arrivals and is more semantically grounded compared to the linear classifiers used in CuMix. Furthermore, our effective text representation through prompt learning is crucial for learning the relationship between image-text pairs, which is evident from the improvement over attribute-based algorithms like BDCZSL. We also observe significant improvements when using Transformer-based text encoders instead of Word2Vec in our algorithm as the text representation for almost all methods. Our network, which uses CLIP as a backbone, achieves significantly better results compared to the raw CLIP model (CLIP-Frozen) and fine-tuned CLIP with contrastive learning (CLIP-Finetune). We attribute the majority of gains to fine-tuning on the DACZSL setting with a CLIP backbone using our factorized features.

Non-uniform DACZSL is more challenging than uniform DACZSL because it requires transferring knowledge from imbalanced and limited sources. As shown in Table~\ref{non-uniform-daczsl-results} and similar to the uniform setting, our method demonstrates clear improvement over all baselines. Additionally, when comparing uniform and non-uniform DACZSL results, our method proves to be robust to randomly removed-domain bias with only minor performance drops, maintaining stable performance in both settings. However, baseline methods exhibit significant declines from uniform DACZSL to non-uniform DACZSL.

In the \iwildcamczsl{}\cite{beery2021iwildcam} benchmark, we apply the Transformer-based text encoder for all baselines, considering the class names in this task often contain obscure words. The results of these experiments, presented in \cref{tab:iwild-cam-results}, indicate that our method significantly outperforms the other baselines in sequentially classifying species. Our method's higher harmonic mean (mH) suggests that our specially designed prompt learning and prototype-based classification efficiently mitigate bias towards seen classes when encountering unseen classes.

\subsection{Ablation Studies}
In this section, we conduct comprehensive analyses of our proposed DIN on the uniform DACZSL setting in the \domainnetczsl{} benchmark. Our method consists of several main components, including the global module, local modules, disentanglement loss, discriminators for domains and tasks, the learnable prompt module, and memory rehearsal. We perform ablation studies on each of these components to evaluate their individual contributions to our approach.

\noindent\textbf{Learnable Prompts.}
CLIP~\cite{radford2021learning} provides a standard prompt template, e.g., ``\texttt{A picture of a [CLASS]}''. However, this fixed template may not be an optimal representation for complex benchmarks. To address this issue, we introduced class-wise learnable prompts (CWP) and performed extensive analyses. First, we assess the experiments with and without prompt learning in Table~\ref{prompt-ablation-daczsl-results}. DIN with CWP shows a noticeable improvement, e.g., around 1.5\%, in harmonic mean accuracy with only a slight sacrifice in BWT around 0.1\%. The slightly lower BWT could be due to less overfitting to seen classes, thus tending to forget more seen information. However, the benefit of unseen information from the learnable prompt is substantial.

Second, we compared Global Learnable Prompts (GLP) over all classes with Class-Wise Prompts (CWP) in~\cref{fig:prompts} and assessed the number $K$ of data points per domain per task needed to learn the prompts. When $K$ is extremely small, GLP tends to outperform CWP. But after fine-tuning on 16 examples per class per domain, we found that CWP outperforms GLP by 0.37\%. Given that $K=16$ does not constitute an overly large set, we choose to use CWP. We also found that using learnable prompts after fine-tuning on increasing samples per class per domain led to a steady increase in harmonic mean accuracy. Moreover, the performance difference between class positions \texttt{mid} and \texttt{end} was marginal. Therefore, our final design choice was based on the simpler \texttt{end} form and class-wise learnable prompt.

\noindent\textbf{Global and Local Modules.} We designed the global network to store domain-invariant and task-invariant knowledge, while the local networks extract task-variant features. We conduct a cumulative ablation study on our method DIN in the scenario with one memory rehearsal. The results, shown in Table~\ref{cumulative-results}, indicate that adding Local Nets can significantly improve LS by 7.53\%, demonstrating a better understanding of seen classes. More importantly, adding Local Nets can facilitate positive knowledge transfer over tasks, e.g., the backward transfer increases from 0.51 to 0.65.

\begin{table}[t]
\centering
\resizebox{\textwidth}{!}{
\begin{tabular}{l|ccc|ccc|ccc|ccc}
\toprule 
Target Domain& \multicolumn{3}{|c|}{Sketch} & \multicolumn{3}{|c|}{Clipart} & \multicolumn{3}{|c|}{Painting} & \multicolumn{3}{c}{Avg.}\\ \cmidrule(l){1-4} \cmidrule(l){5-7} \cmidrule(l){8-10} \cmidrule(l){11-13}
 {Method} & LS$\uparrow$ & mH$\uparrow$ & BWT$\uparrow$ & LS$\uparrow$ & mH$\uparrow$ & BWT$\uparrow$ & LS$\uparrow$ & mH$\uparrow$ & BWT$\uparrow$ & LS$\uparrow$ & mH$\uparrow$ & BWT$\uparrow$\\  
\midrule
{DIN} w/o CWP  & 74.20  & 61.64  & \textbf{1.07 } & ~\textbf{82.15 } & 70.00 &\textbf{ 1.15}  & ~\textbf{77.16 }  & 75.74  & {{\textbf{1.95}}} & \textbf{77.84 } & 69.13  & \textbf{1.39 }\\
{DIN} & \textbf{74.53}  & \textbf{63.78} & 0.92  & {81.31 } & \textbf{71.24 } & 1.12 & {76.00} & \textbf{{76.10}}  & {1.82}  & ~{77.28}  &   \textbf{70.37 } & {1.29}  \\  
\bottomrule\end{tabular}}
\caption{Ablation studies of learnable prompts on \domainnetczsl{} in uniform settings with \texttt{Sketch}, \texttt{Clipart}, and \texttt{Painting} as the target domains, respectively.}
\label{prompt-ablation-daczsl-results}
\vspace{-6mm}
\end{table}

\begin{table}\centering
\begin{minipage}[t]{0.5\textwidth}\small
\resizebox{1.0\textwidth}{!}{%
\begin{tabular}{l|ccc}
\toprule 
 Module & LS$\uparrow$ & mH$\uparrow$ & BWT$\uparrow$\\
\hline
 DIN w/o CWP & 77.84 & 69.13 & 1.39 \\ \hline
 \quad - w/o DD & 76.92 & 68.72 & 0.81\\
 \quad - w/o DT & 75.63 & 67.64 & 1.20\\ \hline
 \quad - w/o DD+DT & 75.58 & 66.92 & 0.79 \\\hline
  \quad - memory & 75.56 & 67.87 & 1.21\\ 
\bottomrule
\end{tabular}  
}
\caption{Ablation studies on memory rehearsal and adversarial training averaging. The results are averaged over all target domains.}\label{advs-results}
\end{minipage}
\hfill
\begin{minipage}[t]{0.45\textwidth}\small
\resizebox{1.0\textwidth}{!}{%
\begin{tabular}{cccc|ccc}
\toprule 
 G & L & Adv & Disen  & LS$\uparrow$ & mH$\uparrow$ & BWT$\uparrow$\\
\hline
 \checkmark  & ~ & ~ & ~ & 64.22 & 64.81 & 0.51 \\ \hline
 \checkmark & \checkmark & ~ & ~ & 71.75 & 65.44 & 0.65\\ \hline
 \checkmark & ~ & \checkmark & ~ & 26.62 & 28.87 & -13.08\\ \hline
 \checkmark & \checkmark & \checkmark & ~ & 76.34 & 68.17 & 0.61\\ \hline
 \checkmark & \checkmark & ~ & \checkmark & 75.58 & 66.92 & 0.79\\ \hline
 \checkmark & \checkmark & \checkmark & \checkmark & 77.84 & 69.13 & 1.39\\
\bottomrule
\end{tabular}    
}  
\caption{Cumulative ablation studies on DIN. The results are averaged over all target domains.}\label{cumulative-results}
\end{minipage}
\vspace{-4mm}
\end{table}

\noindent\textbf{Adversarial Learning and Disentanglement.} Adversarial learning and disentangled learning signals are crucial for creating a global network that contains domain-invariant and task-invariant information. We conducted an ablation study on the impact of adversarial training and disentanglement loss on our method's performance. Results shown in \cref{cumulative-results} indicate that removing adversarial loss leads to a decrease in LS and mH performance by 2.26\% and 2.21\%, respectively. Conversely, removing the disentanglement loss shows a marginal improvement in performance.

Adding adversarial loss without local networks results in a significant performance drop—37.6\% in LS and 35.94\% in mH. The forgetting rate also severely decreases by 13.59 after adversarial training, suggesting that the features extracted by the global net are truly general, yet not directly applicable for classifying images from different domains and categories sequentially.

We also examined the impact of removing domain and task discriminators separately. Results in \cref{advs-results} demonstrate that accuracy over all metrics is more affected by the task discriminator (LS/mH drops by 2.21\%/1.49\% after removing DT) compared to the domain discriminator (LS/mH drops by 0.92\%/0.41\% after removing DD), while the domain discriminator contributes more to reducing forgetting (BWT decreases by 0.58 after removing DD vs. 0.19 after removing DT).

\noindent\textbf{Memory Rehearsal.} Focusing on optimizing memory usage, we stored only one sample per domain per class for each task. Our results, as seen in \cref{advs-results}, reveal an improvement of 2.26\% in LS and 2.21\% in mH with a single memory rehearsal. Additional ablation studies with more samples are detailed in the supplementary material. We observed steady performance gains when increasing memory from 1 to 4 examples. Beyond this, the benefit of expanding memory rehearsal from 4 to 16 examples is minimal. This outcome suggests that our LocalNets-GlobalNet coherent module effectively retains knowledge from previous tasks, mitigating catastrophic forgetting. The module's efficiency in maintaining high classification accuracy, coupled with our model's ability to learn efficiently from a small number of rehearsal samples, underscores its effectiveness.

\subsection{Global Network Latent Feature Analysis}
\begin{wrapfigure}{r}{0.49\textwidth}
\centering
\vspace{-8mm}
\includegraphics[width=0.48\textwidth, trim=35 32 0 0, clip]{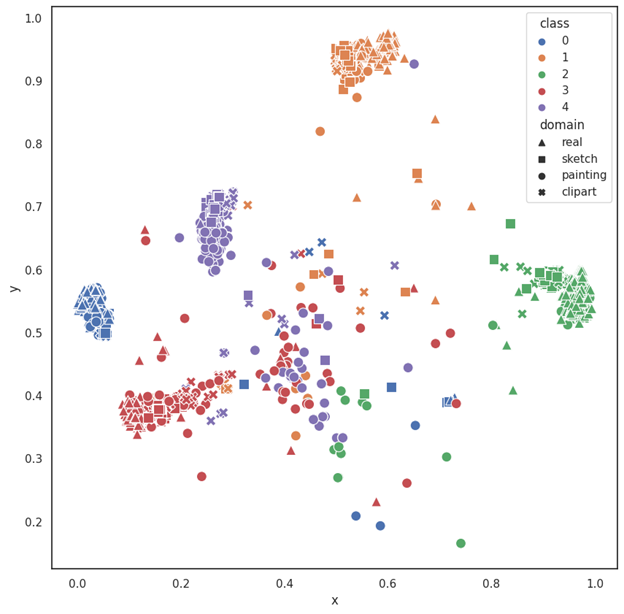}
\caption{t-SNE visualization of latent features from the global network after training on DomainNet-CZSL task 1, highlighting 5 classes.}
\label{fig:global_0}
\vspace{-2mm}
\end{wrapfigure}

During the first task of DomainNet-CZSL, we delved into the latent feature extraction capabilities of the global network, illustrated in Figure~\ref{fig:global_0}. The visualization employs distinct colors and shapes to denote individual classes and their domain-specific features. A notable observation is the close clustering of features within the same class, which distinctly separates from features of other classes. This pattern demonstrates the network's adeptness at identifying class-specific attributes. Moreover, the merging of features from diverse domains within the same class highlights the network’s competence in capturing domain-invariant characteristics. This ability to distill domain-invariant features is crucial for the network's ability to generalize to new, unseen domains in CZSL scenarios. Our analysis confirms that the global network is proficient at pinpointing and extracting both class-specific and domain-invariant features, which are essential for the success of CZSL.

\section{Conclusion}
In this paper, we introduced the Domain-Aware Continual Zero-Shot Learning (DACZSL) task, which addresses the challenge of recognizing images from unseen categories and domains in a lifelong learning setting. Through the example of species classification in camera-trap images, we demonstrated the relevance of this problem. We proposed two new benchmarks for the domain-aware continual zero-shot learning task, \domainnetczsl{} and \iwildcamczsl{}, and presented a novel contrastive prototypical knowledge disentangled representation network, DIN. DIN utilizes a more representative text feature extractor and extracts factorized features. Our experiments showed that improved text representation enhances performance on extreme out-of-distribution images, and that disentangling domain-invariant and task-invariant information is crucial for achieving promising results. We hope that this work will inspire future research into knowledge transfer in out-of-distribution settings and lead to better vision-language models capable of continually understanding the relationships between seen and unseen classes. Additionally, we believe that the models developed in this setting can serve as a foundation for studying animals in the real world, where online distribution shifts and unseen classes are naturally encountered.

%
%
\bibliographystyle{splncs04}
\bibliography{egbib}
\newpage
\clearpage
\appendix
\section{Training Details and Algorithms}
\subsection{Architecture and Training Details}
Our proposed method, DIN, incorporate several key components, including a class-wise learnable prompt, a global network, local networks, a domain discriminator, and a task discriminator. In the following sections, we will provide more detailed explanations of the training and optimization strategies for each module. Additionally, we have included the code for our method as a supplementary resource.

\paragraph{Class-Wise Learnable Prompt Module.} This is a key component of our proposed method. Previous work such as CLIP~\cite{radford2021learning} used a naive prompt in the form of "An image of a [\texttt{CLASS}]," which may not capture class-specific information across different datasets. Zhou et al.\cite{zhou2021learning} demonstrated that a more tailored prompt for a specific dataset could significantly improve classification accuracy on unseen data. Building on this insight, we propose a domain-invariant class-wise learnable prompt module. The general form of the learnable prompt is described in Eqn. 4 of the main paper. The initialization methods for the General Learnable Prompt (GLP) and Class-Wise Prompt (CWP) differ. Our prompt module is represented as a tensor of shape (\# classes, prompt length, dim per prompt), with GLP initializing each class tensor to be identical, while CWP initializes each class tensor uniquely. We train the prompt module by accessing $K$-training examples per class and domain, with a learning rate of 0.002 and $K$ set to 16 by default. For experiments with different values of $K$, we train for 200 epochs for 16/8, 100 epochs for 4/2, and 50 epochs for 1. We use a batch size of 256 for all experiments, including baselines and methods, along with their variants\cite{radford2021learning, zhou2021learning}.

\paragraph{Global and Local Coherent Module.} This module is another important component of our proposed method. We use ResNet50~\cite{he2016deep} pre-trained from CLIP~\cite{radford2021learning} as the visual encoder for both the global and local networks. However, the global network is designed to store domain and task-invariant information, while the local networks are designed to process task-variant information from different source domains. We adopt the Transformer~\cite{vaswani2017attention} as the text encoder, pre-trained from CLIP. The learning rate for these modules is set to 5e-7. We use the AdamW optimizer~\cite{loshchilov2017decoupled} and set the warmup length to 5 and weight decay to 0.02. Additionally, we use the cosine learning scheduler to adapt the learning rate. We set the training epochs for all our modules to 25.

\paragraph{Domain and Task Discriminators.} They are also crucial components of our proposed method. For both discriminators, we use 3-layer MLPs incorporated with LeakyReLU activation. The learning rate is set to 0.001, and weight decay is set to 0.01. We use SGD for optimization, and the latent dimension of the discriminators is set to 1024. Similar to ACL~\cite{ebrahimi2020adversarial}, we add the Gradient Reversal layer~\cite{ganin2015unsupervised} before feeding the generated latent features by the global network to the domain and task-specific discriminators.

\paragraph{CuMix/CNZSL and Their Variants.} We conduct ablation experiments to validate critical hyperparameters and mainly use the default best hyperparameters provided. For CuMix~\cite{mancini2020towards}, we set the hyperparameters of the best-performing model on DACZSL and related experiments to have an image mixup weight of 0.001, feature mixup weight of 0.5, mixup step of 2, and mixup beta of 2. The backbone learning rate is set to 0.0001, and the mapping network learning rate is set to 0.001, with weight decay of 0.00005. The training epoch is set to 25. For CNZSL~\cite{skorokhodov2021class}, we adopt the same design and apply task-specific class normalization. For CuMix, CNZSL, and their variants using Transformer, which is pre-trained from CLIP~\cite{radford2021learning}, to extract the text features, we set the epoch equal to 50. We notice that for CNZSL and its related experiments, we set the batch size to 512, while we set the batch size to 360 for CuMix related experiments due to limited memory.

\subsection{iWildCam DACZSL Setting (Complementary)}

We design the \iwildcamczsl{} benchmark using the version of iWildCam provided by the Wilds~\cite{pmlr-v139-koh21a} benchmark. Dataset consists of 181 different animal species and extra class indicating empty scene. The location of the camera traps are regarded as domains. The total amount of samples found in the dataset is 203,029. We divide the dataset into 7 tasks. Each task introduces 26 classes to the model. 
The dataset has a predefined domain split for training and evaluation. The training domain split contains 243 camera traps. The testing domain split consists of 48 different camera traps. Since each class does not have images captured from all the possible domains, DACZSL on top of \iwildcamczsl{} is naturally a non-uniform setting. We consider each as source and target in our DACZSL setting.

\section{DACZSL Complemantary Results}
\subsection{Evaluation Metrics}  
Inspired by standard continual learning literature, we compute the seen accuracy after training at the last task:  

\begin{equation}
    \operatorname{LS} = \frac{1}{T} \sum_{t=1}^{T} \operatorname{Acc}(X_t^d | \mathcal{M}_T), \quad d \in \mathcal{D}_i.
\end{equation}

\noindent where $\mathcal{D}_i$ is the target domain and $\mathcal{M}_t$ is the model trained till  step $t$. $\operatorname{Acc}(X_t^d | \mathcal{M}_T)$ is the classification accuracy on image set $X^d_t$ with the model trained till task $T$. We also compute the mean seen and unseen accuracy over tasks:

\begin{equation}   
    \begin{aligned}
        \operatorname{mS} &= \frac{1}{T} \sum_{t=1}^{T} \sum_{k=1}^t \operatorname{Acc}(X_k^d | \mathcal{M}_t)\\
        \operatorname{mU} &= \frac{1}{T-1} \sum_{t=1}^{T-1} \sum_{k=t+1}^T \operatorname{Acc}(X_k^d | \mathcal{M}_t).\\
    \end{aligned}
\end{equation}

We consider the harmonic mean mH of mS and mU to exam how balance the method is over seen and unseen accuracy. mH is defined as $2 \operatorname{mS} \cdot \operatorname{mU} / (\operatorname{mS} + \operatorname{mU})$.

Besides, we also consider the measure of forgetting rate:

\begin{equation}  
    \operatorname{BWT} = \sum_{t=1}^{T} \operatorname{Acc}(X_t^d | \mathcal{M}_T) - \operatorname{Acc}(X_t^d | \mathcal{M}_t).
\end{equation}
\subsection{Compare with CL Methods} Our proposed DACZSL setting involves zero-shot learning (ZSL), continual learning (CL), and domain generalization (DG), so we consider existing works with different combinations of them. Therefore, we adopt the following baselines: CuMix (DG+ZSL) and CNZSL (CL+ZSL). Unfortunately, we can not find open-sourced existing works on CL+DG. We believe that exploring CL papers in our setting improves our experiments section. So we attached the results of EWC~\cite{kirkpatrick2017overcoming} and MAS~\cite{aljundi2018memory} in Tab.~\ref{tab:1}. The results show that our proposed method performs significantly better than EWC and MAS.

\begin{figure}[t]
\centering
\begin{minipage}[b]{0.42\textwidth}
\centering\small
\resizebox{\textwidth}{!}{%
\begin{tabular}{c|ccccc}
\toprule 
Method & LS$\uparrow$ & mS$\uparrow$ & mU$\uparrow$ & mH$\uparrow$ & BWT$\uparrow$\\
\hline
EWC~\cite{kirkpatrick2017overcoming} & 8.49 & 9.25 & 3.43 & 4.92 & 0.81 \\ 
EWC~\cite{kirkpatrick2017overcoming} + Tf & 50.31 & 51.86 & 26.19 & 34.57 & 0.55\\ \hline
MAS~\cite{aljundi2018memory} & 7.82 & 9.07 & 2.88 & 4.21 & 1.01\\   
MAS~\cite{aljundi2018memory} + Tf & 51.23 & 52.89 & 27.61 & 36.08 & 0.94 \\ \hline
\textbf{DIN (Ours)} & \color{blue}{\textbf{74.20}} & \color{blue}{\textbf{77.26}} & \color{blue}{\textbf{51.28}} & \color{blue}{\textbf{61.64}} & 1.07\\
\bottomrule
\end{tabular}  
}  
\captionof{table}{Comparison with adaptive continual learning methods of noise-reduced DomainNet. + Tf means we use CLIP~\cite{radford2021learning} pre-trained text Transformer. We report the average over all possible target domains.}\label{tab:1}
\end{minipage}
\hfill
\begin{minipage}[b]{0.55\textwidth}
\centering
\resizebox{0.8\textwidth}{!}{%
\centering
\begin{tabular}{ccc|c|ccc}
  
\toprule 
$\mathcal{L}_{\text{disen}}$ & $\mathcal{L}_{\text{adv}}$ & $\mathcal{L}_{\text{cont}}$ & \# in Tab. 5 & LS$\uparrow$ & mH$\uparrow$ & BWT$\uparrow$\\
\hline
\checkmark & & & - & 20.16 & 18.61 & 0.09\\ \hline
& \checkmark & & - & 31.97 & 30.05 & 0.31\\ \hline
& & \checkmark& a2 & 71.75 & 65.44 & 0.65\\ \hline
\checkmark & \checkmark & & - & 34.22 & 31.69 & 1.02\\ \hline
& \checkmark & \checkmark & a4 & 76.34 & 68.17 & 0.61\\ \hline
\checkmark & & \checkmark & a5 & 75.58 & 66.92 & 0.79\\ \hline
\checkmark & \checkmark & \checkmark & a6 & 77.84 & 69.13 & 1.39\\
\bottomrule
\end{tabular}    
}  
\captionof{table}{Ablation study on our proposed DIN with $1$-memory rehearsal. \textit{\# in Tab. 5} represents which experimental setting in the main paper Table 5 is the same as the one listed. We report the average performance by leveraging all target domains.}\label{cumulative-results2}  
\end{minipage}
\vspace{-2mm}
\end{figure}

\subsection{Full Ablation Table}
Let us recap the total loss to optimize our model (main text Sec. 4.3 Eqn. 5)

\begin{equation}
    \mathcal{L}_{\text{DIN}} = \lambda_1 \mathcal{L}_{\text{disen}} + \lambda_2 \mathcal{L}_{\text{adv}} + \lambda_3 \mathcal{L}_{\text{cont}},
\end{equation}

where the hyperparameters $\lambda_1$, $\lambda_2$, and $\lambda_3$ control the influence of different loss components in our model. The disentangled loss ($\mathcal{L}_{\text{disen}}$) encourages the features in the global network to be independent of every task-variant local network, while the adversarial loss ($\mathcal{L}_{\text{adv}}$) promotes task and domain-invariant knowledge in the global network. The contrastive loss ($\mathcal{L}_{\text{cont}}$) is used for class-level prompt learning.

We conducted extensive experiments to determine the optimal values of $\lambda_1$, $\lambda_2$, and $\lambda_3$, and report the results with the best hyper-parameters we were able to achieve. As a complementary analysis to Tab. 5 in the main paper, we present the results of our component analysis in Tab.~\ref{cumulative-results2}. What distinguishes our study from Tab. 5 in the main paper is that we performed an ablation study by removing the contractive loss ($\mathcal{L}_{\text{cont}}$), which demonstrates a significant drop in performance when visual-language pairs are learned without contractive learning. These findings suggest that strong language-guided DACZSL is necessary to better understand the relationships between visual and language domains.

\end{document}